\documentclass[runningheads]{llncs}
\usepackage[final,year=2026]{eccv}
\usepackage{eccvabbrv}
\usepackage{graphicx}
\usepackage{booktabs}
\usepackage{tabularx}
\usepackage{microtype}
\usepackage{mathtools}
\usepackage{xcolor}
\usepackage[breaklinks,colorlinks,citecolor=eccvblue,linkcolor=eccvblue,urlcolor=eccvblue]{hyperref}
\usepackage{placeins}
\usepackage{needspace}
\usepackage{float}
\usepackage{tikz}
\usepackage{wrapfig}
\usetikzlibrary{arrows.meta,calc,positioning,decorations.markings}
\hypersetup{
  pdfauthor={Ankit Grover and Rémi Bourgerie},
  pdftitle={Do Sheaf Neural Networks Use Holonomy? A Measure--Intervene--Control Study},
  pdfkeywords={Geometric deep learning, Sheaf Neural Networks, Cellular sheaves, Geometric inductive biases, Holonomy, Interpretability}
}
\newcommand{\SO}{\mathrm{SO}}
\newcommand{\GraphUniverse}{\textsc{GraphUniverse}}
\newcommand{\dflat}{\Delta_{\mathrm{id}}}
\newcommand{\dconst}{\Delta_{\mathrm{const}}}

\title{Do Sheaf Neural Networks Use Holonomy? A Measure--Intervene--Control Study}
\titlerunning{Do Sheaf Neural Networks Use Holonomy?}
\author{Ankit Grover \and Rémi Bourgerie}
\authorrunning{A. Grover and R. Bourgerie}
\institute{KTH Royal Institute of Technology, Stockholm, Sweden\\
\email{\{agrover,remibo\}@kth.se}}

\begin{document}
\maketitle

\begin{abstract}
Geometric architectures are often motivated by internal mechanisms, but accuracy alone does not show whether predictions use them. In Sheaf Neural Networks (SNNs), edge transports form a connection whose cycle products define holonomy. We ask whether training changes triangle holonomy, whether predictions rely on the learned connection, and whether holonomy drives triangle counting. We use basis-independent loop readouts with identity interventions and shortcut controls. On high-homophily \GraphUniverse{} graphs, triangle counting increases the mean $\SO(2)$ triangle rotation in Neural Sheaf Propagation (NSP) from $0.010$ to $0.388$ radians, while community detection ends at $0.029$ radians. With more data, learned $\SO(2)$--NSP outperforms Identity NSP, and replacing its transports after training increases error further. However, ridge regression is more accurate, diagonal maps improve without continuous rotation, and fixed-degree models develop rotation without improved counting. Thus NSP can learn and rely on a nontrivial connection, but our experiments do not show that triangle holonomy drives its predictions.
\end{abstract}

\section{Introduction}
Geometric architectures are often interpreted through the rules they encode. Architectural guarantees, however, are not mechanistic explanations. An $\mathrm{SE}(3)$-equivariant pose estimator or physics-aware scene-flow model may satisfy a geometric constraint by construction. Accuracy alone cannot reveal its driving cues. Interpretability work makes an analogous distinction between detecting an internal property and testing its behavioral use through a removal intervention~\cite{10.1162/tacl_a_00359}. Reliable geometric learning therefore needs mechanism-level tests, not only benchmark scores.

Sheaf Neural Networks equip a graph with additional local geometry: vector spaces, or \emph{stalks}, on nodes and edges, linked by learned restriction maps at each incidence~\cite{hansen2019spectral,hansen2020sheaf,bodnar2022neural}. These maps induce transports that carry vectors between neighboring node stalks. The collection of transports forms a discrete connection, and their product around a cycle is its holonomy~\cite{barbero2022sheaf}. Triangles are a natural test because message-passing GNNs have known counting limitations and they are the shortest cycles on which a learned connection can exhibit holonomy~\cite{morris2019weisfeiler,chen2020can}.

Prior work finds that non-identity sheaf maps can be dispensable at task level~\cite{hernandez2026necessity} and separates this from fixed-checkpoint reliance~\cite{liu2026learned}, but does not examine their cycle products. Sevetlidis and Pavlidis estimate holonomy by aligning embeddings along closed paths of nearby inputs; we instead compose transports explicitly defined on graph edges~\cite{sevetlidis2026gaugeinvariant}. Higher-order networks take a different route: they represent each triangle explicitly as its own rank-2 cell, rather than leaving it implicit as a three-edge cycle~\cite{hajij2025copresheaf}. Whether graph-based SNN predictions use their learned triangle-loop geometry therefore remains unresolved.

\Needspace{15\baselineskip}
\begin{wrapfigure}[15]{r}{0.34\linewidth}
\centering
\resizebox{0.94\linewidth}{!}{\resizebox{\linewidth}{!}{%
\begin{tikzpicture}[
  >=Latex,
  font=\scriptsize,
  base vertex/.style={circle, fill=black!55, minimum size=3.6pt,
                      inner sep=0pt},
  stalk/.style={circle, draw=eccvblue, line width=1.1pt,
                fill=eccvblue!7, minimum size=9.5mm, inner sep=1pt},
  edge stalk/.style={circle, draw=none, fill=eccvblue,
                     minimum size=2.8pt, inner sep=0pt},
  edge name/.style={font=\tiny, text=black!65, inner sep=0.4pt},
  graph/.style={draw=black!20, line width=0.9pt},
  attachment/.style={draw=black!32, line width=0.8pt,
                     dash pattern=on 0pt off 2.1pt, line cap=round},
  transport/.style={draw=eccvblue, line width=1.35pt,
                    -{Latex[length=4.5pt,width=3.5pt]}},
  transport label/.style={font=\scriptsize, fill=white, inner sep=1.4pt},
  vector/.style={line width=1.15pt,
                 -{Latex[length=4.2pt,width=3.2pt]}}
]
  \useasboundingbox (-2.20,-1.20) rectangle (2.10,1.98);

  \node[base vertex] (b0) at (-0.76,-0.02) {};
  \node[base vertex] (b1) at ( 0.76,-0.02) {};
  \node[base vertex] (b2) at ( 0.00, 0.67) {};

  \node[stalk] (v0) at (-1.35,-0.63) {$\mathcal F(v_0)$};
  \node[stalk] (v1) at ( 1.35,-0.63) {$\mathcal F(v_1)$};
  \node[stalk] (v2) at ( 0.00, 1.38) {$\mathcal F(v_2)$};

  \draw[graph] (b0)--(b1);
  \draw[graph] (b1)--(b2);
  \draw[graph] (b2)--(b0);

  \node[edge stalk] (e1) at ($(b0)!0.5!(b1)$) {};
  \node[edge stalk] (e2) at ($(b1)!0.5!(b2)$) {};
  \node[edge stalk] (e3) at ($(b2)!0.5!(b0)$) {};
  \node[edge name, below=2.2pt] at (e1) {$e_1$};
  \node[edge name, right=2.2pt] at (e2) {$e_2$};
  \node[edge name, left=2.2pt] at (e3) {$e_3$};

  \draw[attachment] (b0)--(v0);
  \draw[attachment] (b1)--(v1);
  \draw[attachment] (b2)--(v2);

  \draw[transport] (v0.south east) to[bend right=12]
    node[transport label, above=1pt] {$T_{0\to1}$} (v1.south west);
  \draw[transport] (v1.north east) to[bend right=13]
    node[transport label, right=1pt] {$T_{1\to2}$} (v2.south east);
  \draw[transport, shorten >=1.4pt] (v2.south west) to[bend right=13]
    node[transport label, left=1pt] {$T_{2\to0}$} (v0.north west);

  \coordinate (o) at (-1.96,0.69);
  \fill[black!45] (o) circle[radius=1.15pt];
  \draw[vector, black!45] (o) -- ++(0.03,0.64)
    node[font=\scriptsize, black!60, above] {$x$};
  \draw[vector, eccvblue] (o) -- ++(0.48,0.45)
    node[font=\scriptsize, eccvblue, above right=-1pt] {$H_Cx$};
  \draw[eccvblue, line width=0.85pt,
        -{Latex[length=3.2pt,width=2.5pt]}]
    ($(o)+(0.02,0.31)$)
    arc[start angle=87,end angle=43,radius=0.31]
    node[pos=0.55, above right=-1pt, font=\scriptsize] {$\theta$};

\end{tikzpicture}%
}}
\captionsetup{justification=raggedright,singlelinecheck=false}
\caption{\textbf{Discrete triangle holonomy.} Edge stalks (blue) receive maps from incident node stalks. Loop transport sends $x$ to $H_Cx$, producing mismatch $\theta$.}
\label{fig:loop-schematic}
\end{wrapfigure}
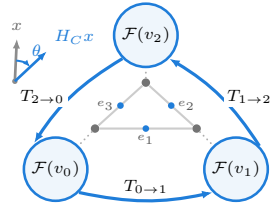
We ask three questions: How do triangle-loop products change during training across tasks? Do predictions rely on the complete learned connection at a fixed checkpoint? After controlling for predictive shortcuts, is there evidence that triangle holonomy is used for triangle counting? Our experiments use \GraphUniverse{}, a synthetic graph generator based on a degree-corrected stochastic block model (DC-SBM). In this graph family, triangle counts can covary with graph size, degree heterogeneity, and community structure~\cite{vanlangendonck2026graphuniverse}. Models may exploit these predictive shortcuts~\cite{geirhos2020shortcut}. We therefore compare three NSP connection conditions and two non-neural models. We also train an SCCN with explicit triangle simplices and generate fixed-degree graphs.

Using TopoBench~\cite{telyatnikov2025topobench}, we \emph{measure} trained triangle-loop products, \emph{intervene} on the complete learned connection at a fixed checkpoint, and use \emph{controls} to assess evidence for triangle-specific computation.

\section{Loop Measurement and Experimental Design}
Each layer learns feature-conditioned restriction maps whose products give transports $T_{u\to v}^{(\ell)}$. For a cycle $C=(v_0,\ldots,v_{k-1},v_0)$, a local vector $x\in\mathcal F(v_0)\simeq\mathbb R^2$ is acted on by
\begin{equation}
x\mapsto H_C^{(\ell)}x,\qquad
H_C^{(\ell)}=T_{v_{k-1}\to v_0}^{(\ell)}\cdots T_{v_0\to v_1}^{(\ell)}.
\label{eq:loop}
\end{equation}
Figure~\ref{fig:loop-schematic} illustrates the closed-loop action. We measure this operator, not one observed activation. All readouts use the final-layer loop product $H_C^{(2)}$, computed from layer-1 representations. Initialization and trained measurements therefore use the same layer (Appendix~\ref{app:transports}). Under independent orthonormal changes of stalk basis, $H_C^{(2)}$ changes only by conjugation. The readouts are therefore basis independent (Appendix~\ref{app:basis-independence}).
For $H=QP$ in $d=2$, the readouts are $\operatorname{twist}(H)=\arccos(\operatorname{tr}Q/2)$, $\operatorname{gain}(H)=|\det H|$, and $\operatorname{flip}(H)=\operatorname{sign}(\det H)$. These measure rotation magnitude, stalk-space area scaling, and orientation reversal.

All three readouts are needed. $\SO(2)$ maps use rotation while fixing gain to one and flip to $+1$. Diagonal maps lack continuous rotation but can change scaling and orientation. General maps combine rotation with anisotropic scaling.

\paragraph{Models.}
We use two-layer Neural Sheaf Propagation (NSP) as the primary model~\cite{suk2022surfing} and repeat the triangle-loop measurements with Neural Sheaf Diffusion (NSD)~\cite{bodnar2022neural}. As an explicit-triangle control, we train a two-layer Simplicial Complex Convolutional Network (SCCN)~\cite{yang2022efficient} on clique-complex lifts, where every graph triangle is a rank-2 simplex (Appendix~\ref{app:sccn-control}). Sheaf models use $d=2$ with diagonal, $\SO(2)$, or general restrictions.

\paragraph{Connection control and intervention.}
Learned $\SO(2)$--NSP trains its transports jointly with all other weights and serves as the reference. Identity NSP is the training-time control: it fixes every transport to $I_d$ while training the remaining weights. Comparing their test errors asks whether learning non-identity transports helps. Identity-replaced NSP is the test-time intervention: it replaces the learned transports with $I_d$ while leaving all other weights unchanged. The error change asks whether the trained model relies on its transports. Because this intervention replaces the complete connection, it does not isolate triangle holonomy (Appendix~\ref{app:connection-controls}).

\paragraph{Tasks, regimes, and evaluation.}
On \GraphUniverse{}, we compare triangle counting with community detection, where node features and homophilous neighborhoods provide predictive cues that do not depend on triangle loops. This comparison uses 30 training and 8 test graphs across 10 splits. A three-seed sweep repeats the connection comparisons with 30--3000 training graphs. A fixed-degree control uses random 6-regular graphs, holding graph size and every node degree constant while triangle counts vary (Appendix~\ref{app:fixed-degree-setup}). For evaluation, triangle counting uses test MSE on the standardized $\log(1+y)$ target computed from training statistics; community detection uses accuracy. Ridge regression tests graph-size and degree shortcuts without access to adjacency or enumerated triangles (Appendix~\ref{app:baselines}, Eq.~\ref{eq:ridge}). We also report $\dconst=\mathrm{MSE}_{\mathrm{model}}-\mathrm{MSE}_{\mathrm{mean}}$, where negative values indicate improvement.

\section{Results}
\begin{figure}[!t]
\centering
\includegraphics[width=0.54\linewidth]{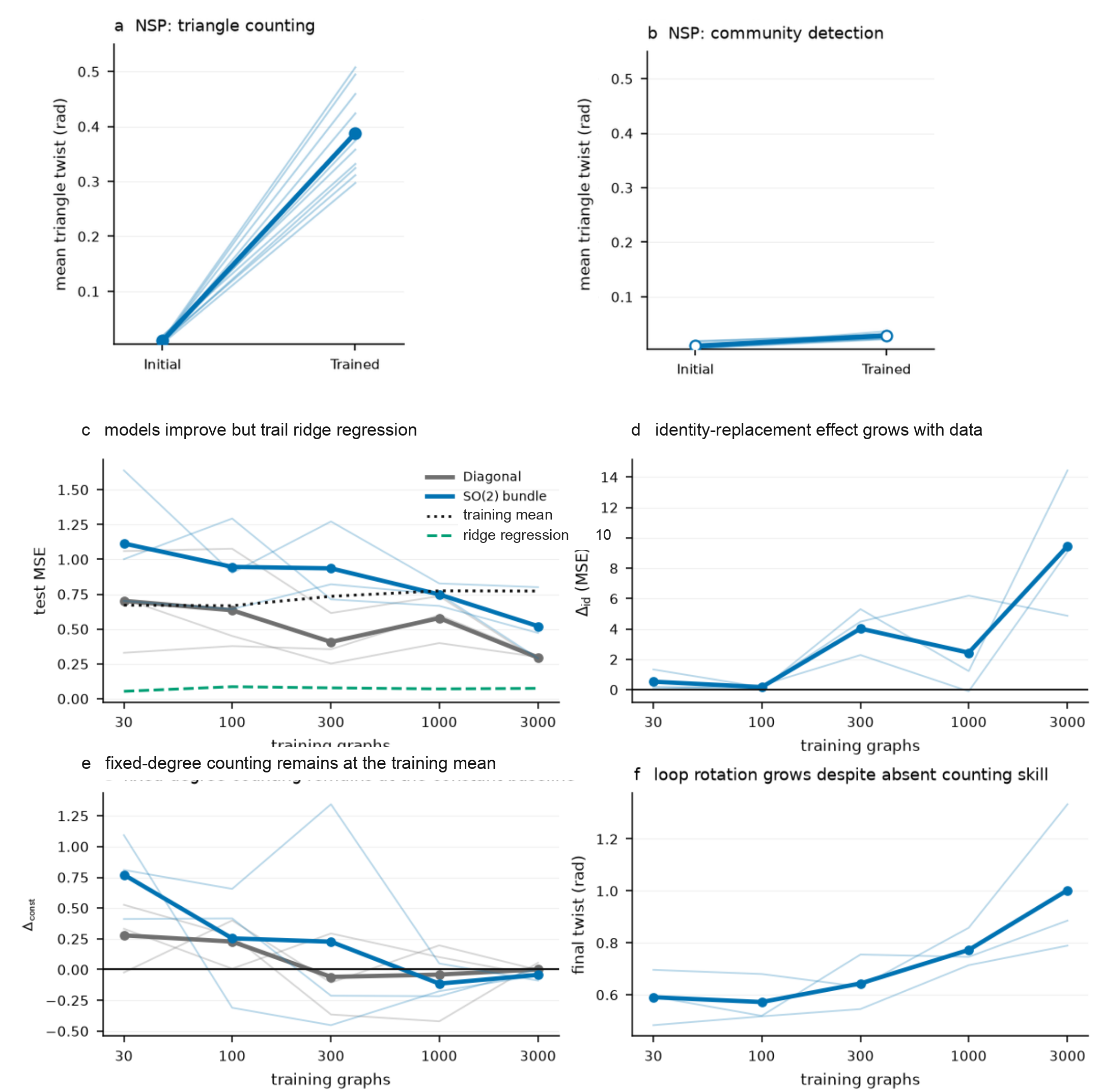}
\captionsetup{justification=raggedright,singlelinecheck=false,skip=2pt}
\caption{\textbf{Measure--intervene--control results in NSP.} Thin lines show runs and thick lines show means. Panels (c--f) use three seeds. (a--b) Counting increases final-layer twist, while community detection remains near identity. With more data, (c) MSE falls but remains above ridge regression, and (d) the identity-replacement effect grows. On random 6-regular graphs, (e) performance remains near the training mean, while (f) twist approaches one radian.}
\label{fig:main-results}
\end{figure}

\paragraph{Training on triangle counting changes the $\SO(2)$ loop products.}
For NSP, triangle counting increases triangle-weighted mean loop rotation from $0.010$ to $0.388\pm0.078$ rad (Appendix~\ref{app:aggregation}). The community-detection comparison ends at $0.029\pm0.005$ rad (Fig.~\ref{fig:main-results}a--b). The increase is similar for NSP and NSD: $0.378\pm0.078$ and $0.405\pm0.075$ rad. For community detection, seven NSD runs stay near identity. The remaining three show high rotation and lower accuracy. We therefore use NSP for the main task comparison and report the NSD results descriptively (Appendix~\ref{app:nsd}).

\smallskip

\paragraph{The setting admits multiple predictive routes.}
At $N_{\mathrm{train}}=30$, ridge regression and SCCN reach MSE $0.110$ and $0.121$. The best graph-based SNN reaches $0.71$. The sampled \GraphUniverse{} regime therefore contains shortcuts based on graph size and degree statistics. It also rewards explicit access to triangles. This does not show that the SNNs use either source.

\paragraph{Learning and retaining the connection both matter.}
Learned $\SO(2)$--NSP has lower mean MSE than Identity NSP throughout the sweep. The separation is clear from $N_{\mathrm{train}}=300$ onward. Replacing the learned connection after training raises the error further. The same pattern appears on 200 new test graphs. Identity NSP tests whether learning the connection helps. Identity-replaced NSP tests whether a trained checkpoint relies on it. Neither result isolates triangle holonomy (Fig.~\ref{fig:main-results}c--d and Appendix~\ref{app:connection-sweep}--\ref{app:bigtest}).

\paragraph{Increasing rotation does not establish triangle-specific computation.}
SCCN has lower mean MSE than each NSP condition (Appendix~\ref{app:sccn-results}). Ridge regression is also highly accurate. Diagonal maps improve even though they cannot express continuous rotations. On fixed-degree graphs, bundle rotation approaches one radian, but performance remains near the training mean (Fig.~\ref{fig:main-results}e--f and Appendix~\ref{app:fixed-degree-results}). These results show that rotation is not sufficient for better counting. They do not identify rotation as the mechanism used by learned $\SO(2)$--NSP.

\enlargethispage{5\baselineskip}
\section{Conclusion}
In the larger-data regime, learned connections have training-time value and fixed-checkpoint reliance. However, ridge regression, diagonal maps, and fixed-degree graphs provide no evidence that triangle holonomy drives predictions. Our basis-independent loop readouts may provide a useful template for testing geometric biases. Limitations appear in Appendix~\ref{app:limitations}.
\clearpage
\bibliographystyle{splncs04}
\bibliography{references}
\clearpage

\setcounter{section}{0}
\renewcommand{\thesection}{\Alph{section}}
\renewcommand{\theHsection}{app.\Alph{section}}
\raggedbottom
\renewcommand{\floatpagefraction}{0.75}
\renewcommand{\textfraction}{0.10}
\renewcommand{\topfraction}{0.90}
\renewcommand{\bottomfraction}{0.80}
\setcounter{topnumber}{3}
\setcounter{bottomnumber}{2}
\setcounter{totalnumber}{5}

\section*{Appendix Contents}
\begingroup
\small
\setlength{\parskip}{1pt}
\newcommand{\appcontentssection}[3]{\noindent\hyperref[#2]{\textbf{#1\quad #3}}\dotfill\pageref{#2}\par}
\newcommand{\appcontentssubsection}[3]{\noindent\hspace*{1.5em}\hyperref[#2]{#1\quad #3}\dotfill\pageref{#2}\par}
\newcommand{\appcontentsunnumbered}[2]{\noindent\hyperref[#1]{\textbf{#2}}\dotfill\pageref{#1}\par}

\appcontentssection{A}{app:math}{Loop-Product Measurement and Validation}
\appcontentssubsection{A.1}{app:transports}{Edge Transports and Loop Products}
\appcontentssubsection{A.2}{app:readouts}{Twist, Gain, and Flip}
\appcontentssubsection{A.3}{app:basis-independence}{Basis Independence}
\appcontentssubsection{A.4}{app:aggregation}{Aggregation over Test Triangles}
\appcontentssubsection{A.5}{app:synchronization}{Local Triangle Loops and Global Connection Consistency}
\appcontentssubsection{A.6}{app:verification}{Validation of Measurement and Identity Replacement}
\medskip

\appcontentssection{B}{app:setup}{Data, Models, and Evaluation Protocol}
\appcontentssubsection{B.1}{app:evaluation-regime}{Graph Regimes and Tasks}
\appcontentssubsection{B.2}{app:generation}{Generating the \GraphUniverse{} Graphs}
\appcontentssubsection{B.3}{app:fixed-degree-setup}{Generating the Fixed-Degree Graphs}
\appcontentssubsection{B.4}{app:models-optimization}{Architectures, Pooling, and Optimization}
\appcontentssubsection{B.5}{app:baselines}{Targets, Metrics, and Non-Neural Models}
\appcontentssubsection{B.6}{app:connection-controls}{Learned, Identity, and Identity-Replaced NSP}
\appcontentssubsection{B.7}{app:sccn-control}{Simplicial Complex Convolutional Network}
\medskip

\appcontentsunnumbered{app:statistics}{Results at a Glance}
\medskip

\appcontentssection{C}{app:geometric-change}{How Training Changes Loop Geometry}
\appcontentssubsection{C.1}{app:nsd}{NSP and NSD Across Tasks}
\appcontentssubsection{C.2}{app:families}{How Restriction Families Change the Readouts}
\medskip

\appcontentssection{D}{app:connection-reliance}{Does Performance Depend on the Learned Connection?}
\appcontentssubsection{D.1}{app:small-replacement}{Identity Replacement with 30 Training Graphs}
\appcontentssubsection{D.2}{app:connection-sweep}{Performance across Training-Set Sizes}
\appcontentssubsection{D.3}{app:bigtest}{Robustness on 200 Test Graphs}
\medskip

\appcontentssection{E}{app:triangle-specificity}{Does Loop Rotation Explain Triangle Counting?}
\appcontentssubsection{E.1}{app:rotation-error}{Loop Rotation versus Counting Error}
\appcontentssubsection{E.2}{app:fixed-degree-results}{Results on Generated Fixed-Degree Graphs}
\appcontentssubsection{E.3}{app:sccn-results}{Comparison with SCCN}
\medskip

\appcontentssection{F}{app:limitations}{Limitations}
\appcontentsunnumbered{app:acknowledgements}{Acknowledgements}
\endgroup
\clearpage

\section{Loop-Product Measurement and Validation}
\label{app:math}

\subsection{Edge Transports and Loop Products}
\label{app:transports}
Let $G=(V,E)$ be an undirected graph. A cellular sheaf assigns a stalk $\mathcal{F}(v)\simeq\mathbb{R}^d$ to every node. It also assigns a restriction map $\mathcal F_{v\unlhd e}^{(\ell)}:\mathcal{F}(v)\to\mathcal{F}(e)$ to each incident node--edge pair at propagation layer $\ell$. For an edge $e=\{u,v\}$, the unnormalized off-diagonal block and derived transport are
\begin{equation}
(L_{\mathcal F}^{(\ell)})_{vu}=-(\mathcal F_{v\unlhd e}^{(\ell)})^{\top}\mathcal F_{u\unlhd e}^{(\ell)},\qquad
T_{u\to v}^{(\ell)}=(\mathcal F_{v\unlhd e}^{(\ell)})^{\top}\mathcal F_{u\unlhd e}^{(\ell)}=-(L_{\mathcal F}^{(\ell)})_{vu}.
\end{equation}
We reserve $\Delta_{\mathcal F}^{(\ell)}=(D_{\mathcal F}^{(\ell)})^{-1/2}L_{\mathcal F}^{(\ell)}(D_{\mathcal F}^{(\ell)})^{-1/2}$ for the normalized Laplacian. For orthogonal restriction maps, transpose equals inverse and the transports form a discrete $O(d)$ connection. TopoBench's two-dimensional bundle parameterization uses the Cayley transform and produces proper rotations in $\SO(2)$.

For a cycle $C=(v_0,v_1,\ldots,v_{k-1},v_0)$, the layer-$\ell$ loop product based at $v_0$ is
\begin{equation}
H_C^{(\ell)}=T_{v_{k-1}\to v_0}^{(\ell)}\cdots T_{v_1\to v_2}^{(\ell)}T_{v_0\to v_1}^{(\ell)}.
\end{equation}
It acts on $x\in\mathcal F(v_0)$ by $x\mapsto H_C^{(\ell)}x$. The measurement records the linear operator induced by the learned connection. It does not trace one observed hidden activation around the graph. Each model layer has its own feature-conditioned sheaf learner. All reported readouts use the final-layer product $H_C^{(2)}$. Its transports are computed from representations already processed by layer 1. Initialization and trained-model measurements both use layer 2. Identity replacement changes transports in both layers. It therefore tests the complete learned connection rather than only the measured final layer.

For isometric transports, $H_C^{(\ell)}$ is the holonomy of the discrete connection. For diagonal and general restrictions, we use the more neutral term \emph{loop product}.

\subsection{Twist, Gain, and Flip}
\label{app:readouts}
Let $H=QP$ be the polar decomposition, with $Q$ orthogonal and $P\succeq0$. Because all experiments use $d=2$, we compute
\begin{align}
\operatorname{twist}(H)&=\arccos\left(\frac{\operatorname{tr}(Q)}{2}\right),\nonumber\\
\operatorname{gain}(H)&=|\det H|,\qquad
\operatorname{flip}(H)=\operatorname{sign}(\det H).
\end{align}
The main text describes these as loop-rotation magnitude, area factor, and orientation sign. In general dimension, the determinant is a volume multiplier. Here $|\det H|$ specifically multiplies two-dimensional stalk-space area. Gain below one means that the loop product contracts area. It does not mean that the graph shrinks. It also does not require every vector norm to decrease. One singular direction may expand while the other contracts more strongly.

The interpretation depends on the restriction family.
\begin{itemize}
\item \textbf{$\SO(2)$ bundle maps.} $P=I_d$, gain is exactly one, flip is $+1$, and twist is the magnitude of the ordinary rotation angle in $[0,\pi]$. This is the setting used for the central rotation claims.
\item \textbf{Diagonal maps.} Loop products remain diagonal. They have no continuous rotation degree of freedom. The orientation sign is the product of edge signs and corresponds to $\mathbb{Z}_2$ frustration in signed-graph balance theory~\cite{cartwright1956structural}.
\item \textbf{General maps.} The polar factor separates rotation from net area change, but its rotation estimate becomes unstable when $H$ is nearly singular.
\end{itemize}
These differences are why we do not combine the three readouts into one generic geometry score.

\subsection{Basis Independence}
\label{app:basis-independence}
Let $G_v\in O(d)$ be an independent orthogonal change of coordinates in each node stalk. Restrictions and transports transform as
\begin{equation}
\mathcal F_{v\unlhd e}^{(\ell)}\mapsto \mathcal F_{v\unlhd e}^{(\ell)}G_v^\top,
\qquad
T_{u\to v}^{(\ell)}\mapsto G_vT_{u\to v}^{(\ell)}G_u^\top.
\end{equation}
Substitution into the loop product gives
\begin{align}
(H_C^{(\ell)})'&=(G_{v_0}T_{v_{k-1}\to v_0}^{(\ell)}G_{v_{k-1}}^\top)\cdots
(G_{v_1}T_{v_0\to v_1}^{(\ell)}G_{v_0}^\top)\\
&=G_{v_0}H_C^{(\ell)}G_{v_0}^\top,
\end{align}
because every interior factor $G_v^\top G_v$ cancels. Thus a loop product changes by conjugation only. Gain and flip are invariant under orthogonal conjugation. For nonsingular $H$, the polar factor is unique and transforms as $Q\mapsto GQG^\top$, so twist is invariant as well. The reported quantities therefore do not depend on the orthonormal coordinate bases chosen in the stalks.

\subsection{Aggregation over Test Triangles}
\label{app:aggregation}
All reported loop summaries are triangle-weighted. If $\mathcal C_3(G)$ is the set of triangles in test graph $G$, the reported mean rotation is
\begin{equation}
\bar\theta_{\mathrm{tri}}=
\frac{\sum_{G\in\mathcal D_{\mathrm{test}}}\sum_{C\in\mathcal C_3(G)}\operatorname{twist}(H_C^{(2)})}
{\sum_{G\in\mathcal D_{\mathrm{test}}}|\mathcal C_3(G)|}.
\end{equation}
Every enumerated triangle contributes equally. Graphs with more triangles therefore contribute more loop observations. Per-graph loop means were not retained. We therefore do not report graph-balanced reweighting. This convention affects the interpretation of the mean. It does not affect the layer, basis-independence, or identity-replacement definitions.

\subsection{Local Triangle Loops and Global Connection Consistency}
\label{app:synchronization}
For $\SO(2)$ transports, global consistency means that a coordinate frame can be assigned to every node. The relative frames at an edge must explain its transport. Such a connection has identity transport around every closed cycle. A nonzero triangle-loop rotation is therefore a local certificate of inconsistency on that triangle.

Singer and Wu use the connection Laplacian to study graph-wide synchronization and frustration~\cite{singer2012vector}. In contrast, we compose learned transports around individual triangles and measure their local loop products. Our measurement does not estimate global spectral frustration or consistency on a longer cycle basis. The two views are complementary rather than interchangeable.

\subsection{Validation of Measurement and Identity Replacement}
\label{app:verification}
Section~\ref{app:basis-independence} proves the invariance property mathematically. The checks below test its numerical implementation. They also verify that bundle loop products remain in $\SO(2)$. A perturbation test measures the stability of bundle rotation. A final check confirms that identity replacement changes the intended transports. These tests validate the measurement and intervention code. They are not additional learning experiments. Reporting them makes the procedure reproducible for future studies that use the same readouts.

\begin{table}[H]
\centering
\scriptsize
\setlength{\tabcolsep}{4pt}
\caption{Numerical checks on the loop measurement and identity-replacement implementation.}
\label{tab:validation}
\begin{tabularx}{\linewidth}{>{\raggedright\arraybackslash}p{2.8cm}>{\raggedright\arraybackslash}p{4.4cm}>{\raggedright\arraybackslash}X}
\toprule
Check & What it establishes & Result\\
\midrule
Basis change & The code agrees with the conjugation-invariance derivation. & Maximum readout change $2.8\times10^{-15}$ across 1,000 random orthogonal basis changes.\\
$\SO(2)$ constraint & Cayley-parameterized transports and their loop products remain proper rotations. & $|\det H|=1.000$ for measured bundle loops.\\
Perturbation stability & Bundle twist is numerically stable at the scale used in the experiments. & Perturbations of size $10^{-4}$ change rotation by $6.86\times10^{-5}$ rad.\\
Nearly singular maps & The same rotation readout is unreliable for ill-conditioned general loop products. & The perturbation changes general-map rotation by $0.195$ rad in the near-singular case.\\
Identity replacement & Both transport layers are replaced as intended and the resulting bundle loops are identities. & $H_C^{(2)}=I_d$: twist $0$, gain $1$, and zero Frobenius error.\\
\bottomrule
\end{tabularx}
\end{table}

\FloatBarrier
\section{Data, Models, and Evaluation Protocol}
\label{app:setup}

\subsection{Graph Regimes and Tasks}
\label{app:evaluation-regime}
The main experiments use a custom high-homophily \GraphUniverse{} regime rather than the benchmark's standard regime grid. The average-degree range is set to 5--7 so that sampled graphs contain enough triangles for repeated loop measurements. Community detection is a comparison task under the same generator. Node features and homophilous neighborhoods provide direct predictive cues. They do not require closed-loop transport. This task is not a cycle-isolating control. The conclusions remain scoped to this synthetic regime.

The second graph regime uses generated random $6$-regular graphs. It complements \GraphUniverse{} by holding graph size, edge count, and the complete degree sequence fixed while wiring and triangle count vary. This removes the graph-size and degree cues available to ridge regression in the main regime.

\subsection{Generating the \GraphUniverse{} Graphs}
\label{app:generation}
\GraphUniverse{} uses a two-stage degree-corrected stochastic block model (DC-SBM).

\paragraph{Fixed universe.}
A master universe is created once with seed 42. It contains $K=20$ universe-level communities and a $20\times20$ edge-propensity matrix with variance 1.0. Node features form 15-dimensional Gaussian clusters. Community centers have variance 0.2. Within-community features have variance 0.4.

\paragraph{Graph-family sampling.}
For each run, graph-level parameters are sampled from the ranges in Table~\ref{tab:gu-settings}. Conditional on community labels $c_u,c_v$ and Pareto degree factors $\theta_u,\theta_v$, edge probabilities are proportional to
\begin{equation}
B_{c_uc_v}\,\theta_u\theta_v,
\end{equation}
then clipped to $[0,1]$. Graphs are made undirected and connectivity is enforced. Data generation and model initialization are independently resampled across seeds. A data split consists of one generated training set and one separately generated test set associated with seed $s$. All methods compared within a split receive exactly those graphs and the same transformed target.

\begin{table}[H]
\centering
\small
\caption{\GraphUniverse{} settings. Ranges are sampled independently for each graph family.}
\label{tab:gu-settings}
\begin{tabularx}{\linewidth}{>{\raggedright\arraybackslash}p{4.1cm}>{\raggedright\arraybackslash}X}
\toprule
Setting & Value\\
\midrule
Nodes per graph & 100--160\\
Communities per graph & 5--7, drawn from 20 universe communities\\
Average degree & 5.0--7.0\\
Power-law exponent & 2.0--2.5\\
Degree separation & 0.5--1.0\\
Homophily & $[0.9,1.0]$\\
Main split & 30 training and 8 test graphs per seed\\
Larger test & 200 newly generated test graphs for selected settings\\
Training-set sizes & $N_{\mathrm{train}}\in\{30,100,300,1000,3000\}$, with the same 8 test graphs within each seed\\
\bottomrule
\end{tabularx}
\end{table}

\subsection{Generating the Fixed-Degree Graphs}
\label{app:fixed-degree-setup}
The fixed-degree graphs are generated separately from \GraphUniverse{} using NetworkX random regular graphs,
\begin{equation}
G_i^{(s)}\sim\texttt{random\_regular\_graph}(r=6,n=100,
\texttt{seed}=1000s+i).
\end{equation}
Every graph has 100 nodes, 300 edges, and degree six at every node. Only its wiring and triangle count vary. We evaluate two feature conditions. Independent 15-dimensional Gaussian features act as symmetry breakers. Constant all-ones features retain anonymous message passing. The split construction and standardized log-count target in Eq.~\ref{eq:target} otherwise match the \GraphUniverse{} experiment.

\subsection{Architectures, Pooling, and Optimization}
\label{app:models-optimization}
\begin{table}[H]
\centering
\small
\caption{Architecture and training settings.}
\label{tab:model-setup}
\begin{tabularx}{\linewidth}{>{\raggedright\arraybackslash}p{4.1cm}>{\raggedright\arraybackslash}X}
\toprule
Setting & Value\\
\midrule
Neural architectures & NSP~\cite{suk2022surfing}, NSD~\cite{bodnar2022neural}, and SCCN~\cite{yang2022efficient}\\
Sheaf restrictions & diagonal, $\SO(2)$ bundle, and general $2\times2$\\
Sheaf model size & 2 layers, stalk dimension $d=2$, and feature width 32\\
Graph-level readout & mean pooling followed by an MLP\\
Loop measurement & final-layer product $H_C^{(2)}$\\
Optimization & Adam with learning rate $10^{-2}$, batch size 8, and 60 epochs\\
Main task comparison & 30 training graphs and $2$ sheaf dynamics $\times$ $2$ tasks $\times$ $3$ restriction families $\times$ $10$ seeds\\
Training-size sweep & $N_{\mathrm{train}}=30,100,300,1000,3000$, with 3 seeds and fixed test graphs within each seed\\
\bottomrule
\end{tabularx}
\end{table}

For graph-level triangle prediction, mean pooling keeps the representation scale from growing directly with graph size or simplex multiplicity. This is compatible with the standardized $\log(1+y)$ target, which grows much more slowly than the raw count. We report this as the chosen parameterization, not as a separately evaluated pooling result.

\subsection{Targets, Metrics, and Non-Neural Models}
\label{app:baselines}
Community detection predicts the universe-level community identifier for each node, with labels in $\{0,\ldots,19\}$. Triangle counting first computes the raw graph-level count
\begin{equation}
y(G)=\frac{1}{3}\sum_{v\in V}\operatorname{triangles}(v),
\end{equation}
where each triangle is counted at its three vertices. For split $s$, the regression target is
\begin{equation}
z_s(G)=\frac{\log(1+y(G))-\mu_s}{\sigma_s},
\label{eq:target}
\end{equation}
where $\mu_s$ and $\sigma_s$ are computed from the training graphs only. They are the mean and standard deviation of $\log(1+y)$. The raw counts span a wide, right-skewed range. The natural-log transform compresses that range. This prevents triangle-heavy graphs from dominating squared error and reduces target-scale variation. Adding one keeps the target defined when $y=0$. Standardization gives the training target zero mean and unit variance within each split. All reported counting MSEs are errors in standardized log-count space. They are not absolute errors in raw triangle counts. Test graphs never contribute to the transformation.

The two non-neural models use the same training and test graphs and the same target $z_s$ as the neural models.

The \emph{training-mean model} ignores the input graph and always outputs $z_s=0$. This is equivalent to predicting $\mu_s$, the training-set mean of $\log(1+y)$. It is not the mean of the raw triangle counts. Its test MSE is
\begin{equation}
\mathcal L_{\mathrm{mean}}=
\frac{1}{|\mathcal D_{\mathrm{test}}|}
\sum_{G\in\mathcal D_{\mathrm{test}}}z_s(G)^2.
\end{equation}
It is computed separately for every split rather than assumed to equal one. The test graphs need not have the training variance. With eight test graphs, this value varies substantially. In the fixed-degree experiment it ranges from approximately 0.43 to 3.28 across seeds.

\emph{Ridge regression} is an $\ell_2$-regularized linear model. It receives seven graph-level inputs. Two are node count and edge count. The other five are the minimum, maximum, mean, standard deviation, and median of the node degrees. For design matrix $Z\in\mathbb{R}^{m\times7}$ and standardized targets $\mathbf z\in\mathbb{R}^m$, its parameters solve
\begin{equation}
(\hat b,\hat{\boldsymbol\beta})=
\arg\min_{b,\boldsymbol\beta}
\left\|\mathbf z-b\mathbf 1-Z\boldsymbol\beta\right\|_2^2
+\lambda\left\|\boldsymbol\beta\right\|_2^2,
\label{eq:ridge}
\end{equation}
where $b$ is an unpenalized intercept and $\lambda\ge0$ controls regularization. Ridge regression never receives the adjacency matrix, node triples, or enumerated triangles. It asks whether the sampled \GraphUniverse{} family permits accurate prediction from graph size and the marginal degree distribution. Its success does not show that the SNNs use the same information.

\subsection{Learned, Identity, and Identity-Replaced NSP}
\label{app:connection-controls}
We compare three conditions built from the same two-layer NSP propagation architecture.
\begin{itemize}
\item \textbf{Learned $\SO(2)$--NSP} jointly optimizes both layers' Cayley-parameterized transports and all remaining model parameters.
\item \textbf{Identity NSP} fixes every transport to $I_d$ from initialization while optimizing all remaining NSP parameters under the same protocol. It is the NSP implementation of an Identity Sheaf Network~\cite{hernandez2026necessity}.
\item \textbf{Identity-replaced NSP} first trains learned $\SO(2)$--NSP. It then zeroes both layers' Cayley parameters without retraining. Because $\operatorname{Cayley}(0)=I_d$, every transport becomes $I_d$. All remaining weights retain the values learned with a non-identity connection.
\end{itemize}
Identity NSP asks whether allowing a non-identity connection improves the model obtained by training. Identity-replaced NSP instead asks whether a particular trained checkpoint relies on its learned connection. This distinction follows the separation between task-level value and checkpoint reliance~\cite{liu2026learned}. The intervention is broad: it changes the complete two-layer connection and does not selectively remove triangle holonomy.

For the fixed-checkpoint intervention we report
\begin{equation}
\dflat=\mathcal L_{\mathrm{identity\text{-}replaced}}-\mathcal L_{\mathrm{learned}}.
\end{equation}
Positive values mean that replacing the learned connection increases error.

\subsection{Simplicial Complex Convolutional Network}
\label{app:sccn-control}
We implement the two-layer Simplicial Complex Convolutional Network (SCCN) of Yang et al.~\cite{yang2022efficient}. Each input graph is lifted to its two-dimensional clique complex. Nodes are rank-0 simplices, and edges are rank-1 simplices. Every graph triangle becomes one rank-2 simplex. The model propagates across simplex ranks. It then mean-pools the learned rank-2 representations and passes the result through an MLP.

SCCN uses the same \GraphUniverse{} splits and standardized $\log(1+y)$ target as the NSP training-size sweep. It also uses the same three seeds, 60 training epochs, and eight test graphs. Its representation is intentionally different from the graph-based SNNs. The clique-complex construction exposes every target triangle before learning. The SCCN result tests whether the protocol rewards explicit triangle simplices. It is not a representation-matched test of the sheaf connection.

\FloatBarrier
\phantomsection
\label{app:statistics}
\section*{Results at a Glance}
Table~\ref{tab:stats-summary} links each main conclusion to its central evidence and the appendix section containing the full result.

\begin{table}[H]
\centering
\scriptsize
\setlength{\tabcolsep}{4pt}
\caption{Guide to the principal results.}
\label{tab:stats-summary}
\begin{tabularx}{\linewidth}{>{\raggedright\arraybackslash}p{4.0cm}>{\raggedright\arraybackslash}X>{\centering\arraybackslash}p{1.4cm}}
\toprule
Finding & Central evidence & Details\\
\midrule
Counting changes bundle loop geometry & NSP twist increases from $0.010$ to $0.388\pm0.078$ rad. Community detection ends at $0.029\pm0.005$ rad. & \ref{app:nsd}\\
The complete learned connection has training-time value & Learned $\SO(2)$--NSP has lower mean MSE than Identity NSP, with clear separation from $N_{\mathrm{train}}=300$ onward. & \ref{app:connection-sweep}\\
Trained checkpoints rely on the learned connection & Identity-replaced NSP has higher mean MSE. The direction persists on 200 new test graphs. & \ref{app:connection-sweep}, \ref{app:bigtest}\\
Accurate counting is possible in this protocol & SCCN reaches $0.070\pm0.058$ MSE at $N_{\mathrm{train}}=3000$ when triangles are explicit simplices. & \ref{app:sccn-results}\\
Rotation alone does not explain counting & Rotation is weakly associated with error. Diagonal and fixed-degree results separate rotation from counting performance. & \ref{app:rotation-error}, \ref{app:fixed-degree-results}\\
\bottomrule
\end{tabularx}
\end{table}

\FloatBarrier
\section{How Training Changes Loop Geometry}
\label{app:geometric-change}

\subsection{NSP and NSD Across Tasks}
\label{app:nsd}
On triangle counting, the final-layer $\SO(2)$ loop-rotation changes are similar for NSP and NSD. They are $0.378\pm0.078$ and $0.405\pm0.075$ rad, respectively. The difference appears in community detection. Seven of ten NSD runs stay near identity. The other three reach roughly $1.6$ rad and have the lowest accuracies in that condition. We therefore report the NSD task comparison descriptively.

\subsection{How Restriction Families Change the Readouts}
\label{app:families}
\begin{figure}[H]
\centering
\includegraphics[width=0.92\linewidth]{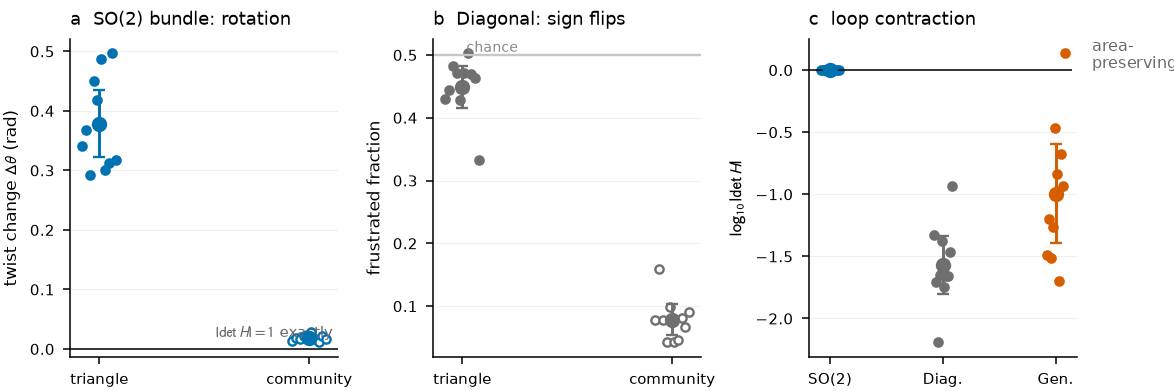}
\caption{\textbf{Restriction families change different loop readouts (NSP).} Triangle counting increases bundle rotation (a). Community detection reduces the fraction of diagonal loops with negative determinant below the $0.5$ chance line (b). Only bundle loops preserve area. Diagonal and general loops contract it (c), where $\log_{10}|\det H|=0$ denotes preservation. Points are seeds. Bars show mean 95\% confidence intervals.}
\label{fig:families}
\end{figure}

Figure~\ref{fig:families} shows the three measured responses. Bundle maps increase final-layer loop rotation while preserving stalk-space area exactly. Diagonal community-detection models reduce the fraction of loops with negative orientation sign, consistent with a more balanced sign pattern. Diagonal and general loop products contract strongly. A single scalar would hide these distinct behaviors and would make the unstable rotation estimate of nearly singular general maps appear more comparable than it is.

\FloatBarrier
\section{Does Performance Depend on the Learned Connection?}
\label{app:connection-reliance}

\subsection{Identity Replacement with 30 Training Graphs}
\label{app:small-replacement}
For triangle counting, identity replacement increases MSE by $0.48\pm0.68$ for NSP and $0.17\pm0.80$ for NSD. Both 95\% confidence intervals include zero. They also include increases large enough to matter. The small-data result is therefore unresolved. For community detection, the mean change is close to zero for both dynamics.

\begin{figure}[H]
\centering
\includegraphics[width=0.92\linewidth]{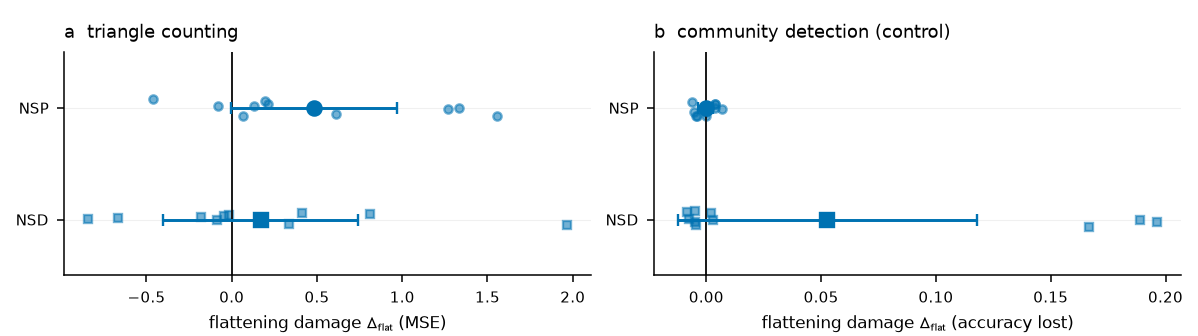}
\caption{\textbf{Identity replacement is unresolved with 30 training graphs.} Zero denotes no change. Positive values mean higher counting MSE (a) or lower community accuracy (b) after replacement. Counting seeds lie on both sides of zero, and both mean 95\% confidence intervals cross it. Community effects are mostly near zero, with three larger NSD accuracy losses. Small marks are seeds. Large marks are means.}
\label{fig:small-replacement}
\end{figure}

\Needspace{12\baselineskip}
\subsection{Performance across Training-Set Sizes}
\label{app:connection-sweep}
Table~\ref{tab:connection-sweep} separates three questions. These are training-time connection value, fixed-checkpoint reliance, and performance with explicit triangle simplices. Learned $\SO(2)$--NSP has lower mean MSE than Identity NSP at every training size. The first two differences are small relative to seed variability. From $N_{\mathrm{train}}=300$ onward, the mean separation is substantial. At $N_{\mathrm{train}}=3000$, learned $\SO(2)$--NSP reaches $0.521\pm0.259$ MSE. Identity NSP reaches $2.724\pm0.549$.

Identity-replaced NSP has higher mean MSE than the learned model throughout the sweep. At $N_{\mathrm{train}}=3000$, it reaches $9.981\pm4.652$. Thus the learned connection has value during training. The resulting checkpoints also depend on retaining it. Neither comparison identifies triangle holonomy as the causal mechanism. Both concern the complete connection.

\begin{table}[H]
\centering
\scriptsize
\setlength{\tabcolsep}{3pt}
\caption{Performance across connection conditions and SCCN. Entries are standardized $\log(1+y)$ test MSE. We report the mean $\pm$ sample standard deviation over three seeds. Lower is better. Identity replacement is applied after training without retraining.}
\label{tab:connection-sweep}
\begin{tabularx}{\linewidth}{>{\centering\arraybackslash}p{1.45cm}>{\centering\arraybackslash}X>{\centering\arraybackslash}X>{\centering\arraybackslash}X>{\centering\arraybackslash}X}
\toprule
$N_{\mathrm{train}}$ & Identity NSP & Learned $\SO(2)$--NSP & Identity-replaced NSP & SCCN\\
\midrule
$30$   & $1.228\pm0.265$ & $1.114\pm0.478$ & $1.644\pm1.167$ & $0.121\pm0.020$\\
$100$  & $1.054\pm0.584$ & $0.946\pm0.327$ & $1.113\pm0.207$ & $0.144\pm0.114$\\
$300$  & $2.512\pm1.607$ & $0.935\pm0.296$ & $4.957\pm1.610$ & $0.091\pm0.075$\\
$1000$ & $2.949\pm1.845$ & $0.748\pm0.080$ & $3.185\pm3.379$ & $0.108\pm0.047$\\
$3000$ & $2.724\pm0.549$ & $0.521\pm0.259$ & $9.981\pm4.652$ & $0.070\pm0.058$\\
\bottomrule
\end{tabularx}
\end{table}

\subsection{Robustness on 200 Test Graphs}
\label{app:bigtest}
The training-size curves use eight test graphs per seed. We selected the pivotal \GraphUniverse{} configurations at $N_{\mathrm{train}}\in\{300,3000\}$. We also selected the fixed-degree configuration at $N_{\mathrm{train}}=3000$. The original checkpoints were not saved. We therefore retrained each configuration with the same split, seed, and protocol. Each retrained model was evaluated on the original eight graphs and on 200 new test graphs. This is a larger-test robustness check, not an additional independent training seed.

The retrained eight-graph metrics closely match the originals. The median absolute difference is zero, and 72\% of runs are within $0.01$. A few $N_{\mathrm{train}}=3000$ runs differ across retraining. On the 200-graph tests, the mean identity-replacement effect remains positive at the selected \GraphUniverse{} settings. The fixed-degree models remain near the training-mean model. The larger test therefore supports the direction of the connection-reliance result. It does not remove the seed variability.

\begin{figure}[H]
\centering
\includegraphics[width=0.92\linewidth]{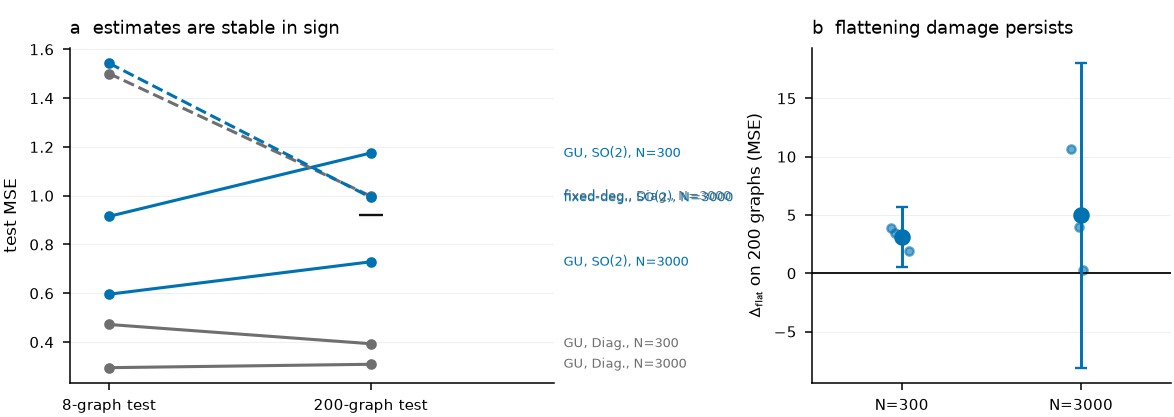}
\caption{\textbf{Larger-test robustness check.} Lines in (a) connect the same retrained model evaluated on 8 and 200 test graphs. They expose shifts caused by the small original test set. On 200 graphs, the mean identity-replacement effect remains positive at $N=300$ and $N=3000$ (b). The latter is highly variable. In (b), small marks are seeds. Large marks and bars are means with 95\% confidence intervals.}
\label{fig:bigtest}
\end{figure}

\FloatBarrier
\section{Does Loop Rotation Explain Triangle Counting?}
\label{app:triangle-specificity}

\subsection{Loop Rotation versus Counting Error}
\label{app:rotation-error}
\begin{figure}[H]
\centering
\includegraphics[width=0.58\linewidth]{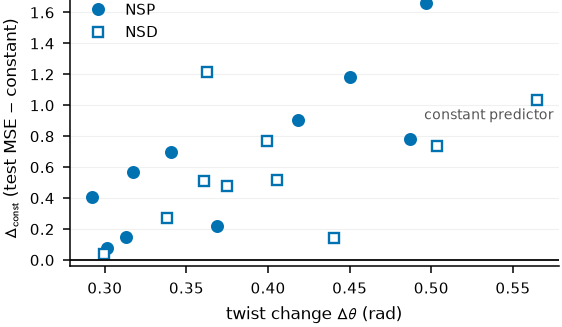}
\caption{\textbf{More loop rotation is not associated with better counting.} Each point is one seed. Circles denote NSP and squares denote NSD. The horizontal axis is the change in final-layer twist. The vertical axis is $\dconst$, where lower values are better. The weak association ($r=-0.18$, $p=0.58$) provides no evidence that larger rotation changes reduce error.}
\label{fig:rotation-error}
\end{figure}

Within the $\SO(2)$ bundle family, loop-rotation change is not negatively associated with triangle-counting error. Across the plotted runs, Pearson $r=-0.18$ with $p=0.58$. Figure~\ref{fig:rotation-error} plots error relative to the training-mean model. We do not pool restriction families in one regression. Family identity dominates that comparison, and continuous rotation is not meaningful for diagonal maps.

\subsection{Results on Generated Fixed-Degree Graphs}
\label{app:fixed-degree-results}
On random $6$-regular graphs with independent Gaussian node features, no restriction family consistently improves on the training-mean model. Bundle maps nevertheless develop substantial loop rotation. With constant node features, all families match the training-mean model to within $|\Delta|\le0.007$. The learned bundle transports remain identities, so identity replacement is vacuous. Random node features can make message passing more expressive than anonymous $1$-WL~\cite{abboud2021surprising}. The observed failure is therefore an empirical optimization and sample-efficiency result, not an expressivity theorem.

\begin{figure}[H]
\centering
\includegraphics[width=0.92\linewidth]{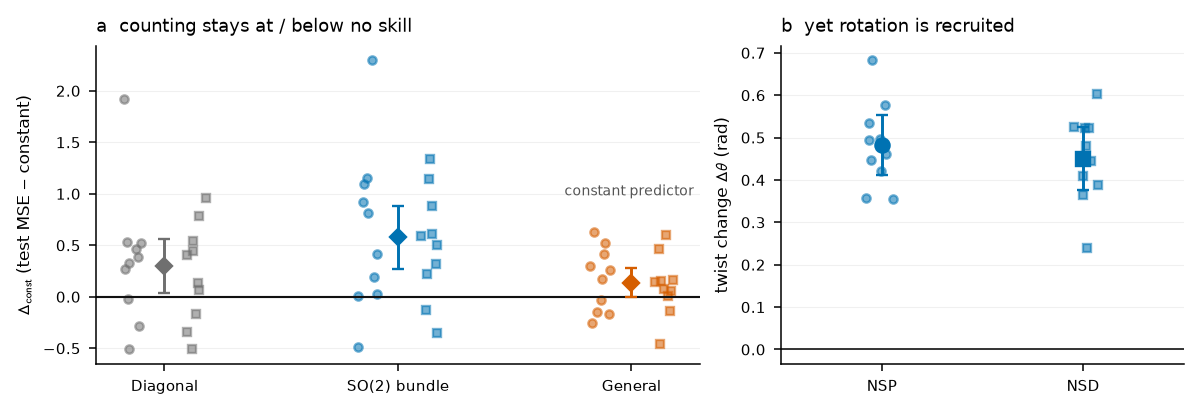}
\caption{\textbf{Fixed degree separates loop rotation from counting skill.} With Gaussian node features, no restriction family consistently falls below the $\dconst=0$ training-mean reference in (a). Nevertheless, bundle twist increases for both NSP and NSD (b). Small marks are seeds. Large marks and bars are means with 95\% confidence intervals.}
\label{fig:regular}
\end{figure}

\subsection{Comparison with SCCN}
\label{app:sccn-results}
SCCN has lower mean MSE than all three NSP conditions at every training size in Table~\ref{tab:connection-sweep}. At $N_{\mathrm{train}}=3000$, it reaches $0.070\pm0.058$. This shows that the protocol rewards explicit access to triangle simplices. It also addresses an alternative explanation: perhaps no trained model could learn the target under the chosen splits and transformed loss.

The result does not make SCCN a representation-matched baseline for the graph-based SNNs. The clique-complex lift enumerates every target triangle before learning. The SNNs receive only nodes and edges and must use edge-based propagation. SCCN therefore establishes learnability with explicit higher-order structure. It does not show that the SNNs should recover the same solution. It also does not show that their learned connection stores triangle counts.

\FloatBarrier
\section{Limitations}
\label{app:limitations}
The experiments cover synthetic high-homophily graphs, two-dimensional stalks, and triangle loops. The training-size study uses three seeds. The main task comparison relies on NSP because three NSD community-detection runs behave differently from the other seven. Rotation from nearly singular general loop products is unreliable and is used only descriptively.

Identity replacement is a broad intervention. It changes both learned transport layers after training. This may disrupt weights that adapted to the original connection. Identity NSP reduces this concern. Learned $\SO(2)$--NSP also has lower mean error when the alternative is trained from initialization. Nevertheless, neither comparison reveals what information the complete connection stores. Neither isolates triangle holonomy. A matched-size connection perturbation or a cycle-consistency objective would separate those questions more directly. Training for a fixed 60 epochs also gives larger datasets more optimizer updates. Data size and optimization effort are therefore not fully separated.

SCCN confirms that explicit triangle structure is useful. Its clique-complex lift is not representation matched to the graph-based SNNs because it enumerates the target motif. Ridge regression further shows that the sampled \GraphUniverse{} regime contains strong graph-size and degree cues. The conclusions may not extend to longer cycles or less structured graph families. They may also differ for higher-order models that are not given the counted motif explicitly.

\phantomsection
\label{app:acknowledgements}
\section*{Acknowledgements}
This work was conducted independently, but benefited from discussions at the 2026 London Geometry and Machine Learning (LOGML) Summer School. We are grateful to its participants and organizers for creating a stimulating environment. We also thank Modal for providing free GPU credits.

\end{document}